\pdfoutput=1

\documentclass[11pt]{article}

 \usepackage[]{acl}

\usepackage{times}
\usepackage{latexsym}
\usepackage[T1]{fontenc}
\usepackage{listings}

\usepackage[utf8]{inputenc}

\usepackage{microtype}

\usepackage{times}
\usepackage{latexsym}

\usepackage{url}
\usepackage{multirow}
\usepackage{microtype}
\usepackage{booktabs}
\usepackage{graphicx}
\usepackage{hyperref}
\usepackage{amsmath}
\usepackage{amsthm}
\usepackage{amsfonts}
\usepackage{tikz}
\usepackage{amsmath}
\usepackage{booktabs}
\usepackage{pifont}
\usepackage{enumitem}
\usepackage{amssymb}
\usepackage{subcaption}
\usepackage{booktabs}
\usepackage{float}
\usepackage[]{xcolor}
\usepackage{times}
\usepackage{footnote}
\usepackage{latexsym}
\usepackage{todonotes}
\usepackage{url}
\usepackage{xargs}
\usepackage{enumitem}
\usepackage{booktabs} 

\newcommand{\Table}[1]{Table~\ref{#1}}

\newcommand{\Figure}[1]{Fig.~\ref{#1}}

\newcommand{\Section}[1]{\S\ref{#1}}
\newcommand{\ignore}[1]{}

\newcommand\ie{\emph{i.e.}}
\newcommand\eg{\emph{e.g.}}

\newcommand{\winogrande}{WinoGrande}
\newcommand{\hellaswag}{HellaSwag}
\newcommand{\socialiqa}{Social IQa}
\newcommand{\piqa}{PIQA}
\newcommand{\pretrained}{pre-trained}

\newenvironment{itemizesquish}{\begin{list}{\labelitemi}{\setlength{\itemsep}{0em}\setlength{\labelwidth}{0.5em}\setlength{\leftmargin}{\labelwidth}\addtolength{\leftmargin}{\labelsep}}}{\end{list}}

\title{A Systematic Investigation of Commonsense Knowledge \\in Large Language Models}

\author{
	Xiang Lorraine Li$^\dagger$ \thanks{\quad Work done during DeepMind internship when Lorraine was a PhD student at UMass Amherst. $\diamondsuit$ Work done at DeepMind} \qquad  
	Adhiguna Kuncoro$^\ddagger$ \qquad 
	Jordan Hoffmann$^{\bigstar  \diamondsuit}$ \qquad \\
	\textbf{Cyprien de Masson d'Autume$^{\blacklozenge  \diamondsuit}$ \qquad}
	\textbf{Phil Blunsom$^{\blacktriangle \clubsuit  \diamondsuit}$} \qquad  
	\textbf{Aida Nematzadeh$^\ddagger$}   \\ \\
$^\dagger$ Allen Institute for Artificial Intelligence \qquad
$^\ddagger$ DeepMind \qquad \\
$^\bigstar$ Inflection AI \qquad 
$^\blacklozenge$ Reka \qquad
$^\blacktriangle$ Cohere \qquad $^\spadesuit$ University of Oxford \qquad
 \\ \\
{ lorrainel@allenai.org \qquad nematzadeh@google.com} \\
}

\date{}

\begin{document}
\maketitle
\begin{abstract}
Language models (LMs) trained on large amounts of data \citep[\eg,][]{gpt3, mt-nlg} have shown impressive performance on many NLP tasks under the zero-shot and few-shot setup. Here we aim to better understand the extent to which such models learn commonsense knowledge --- a critical component of many NLP applications. We conduct a systematic and rigorous zero-shot and few-shot commonsense evaluation of large \pretrained{} LMs, where we: (i) carefully control for the LMs' ability to exploit potential surface cues and annotation artefacts, and (ii) account for variations in performance that arise from factors that are not related to commonsense knowledge. Our findings highlight the limitations of \pretrained{} LMs in acquiring commonsense knowledge without task-specific supervision; furthermore, using larger models or few-shot evaluation are insufficient to achieve human-level commonsense performance. 
\end{abstract}

\section{Introduction}
Common sense --- the implicit knowledge about everyday situation that is shared by humans --- is an important prerequisite for developing general-purpose intelligent systems~\citep[][]{mccarthy1960programs,liu2004commonsense,gunning2018machine}. Intriguingly, recent large language models ~\citep[LMs,][]{gpt3, mt-nlg, rae-etal-2021-gopher} have achieved remarkable performance at various common sense benchmarks \citep[\eg,][]{winogrande,hellaswag,piqa,socialiqa}, even when they are evaluated in a zero-shot or few-shot fashion, \emph{without} explicit commonsense supervision.
We revisit this apparent success, and conduct a rigorous study to better understand the extent to which such \pretrained{} LMs are able to capture commonsense knowledge.

In this work, we focus on zero- and few-shot evaluations of \pretrained{} LMs without commonsense-specific fine-tuning for two reasons: First, we aim to examine if a \pretrained{} LM is able to acquire \emph{general} commonsense knowledge. As \pretrained{} LMs constitute a \emph{foundational} building block of NLP today, any deficiencies in their commonsense understanding can thus adversely manifest in downstream applications \citep{Bommasani2021OnTO}. Fine-tuning the LM would make it hard to disentangle how much of the commonsense knowledge is acquired by the underlying LM, as opposed to the \emph{task-specific} supervision from a benchmark~\citep{yogatama2019learning}. Second, human-annotated commonsense datasets are expensive to collect due to the vast, diverse, and growing nature of commonsense knowledge \citep{elazar2021back}.

\begin{figure}[t]
\centering
\includegraphics[width=0.5\textwidth]{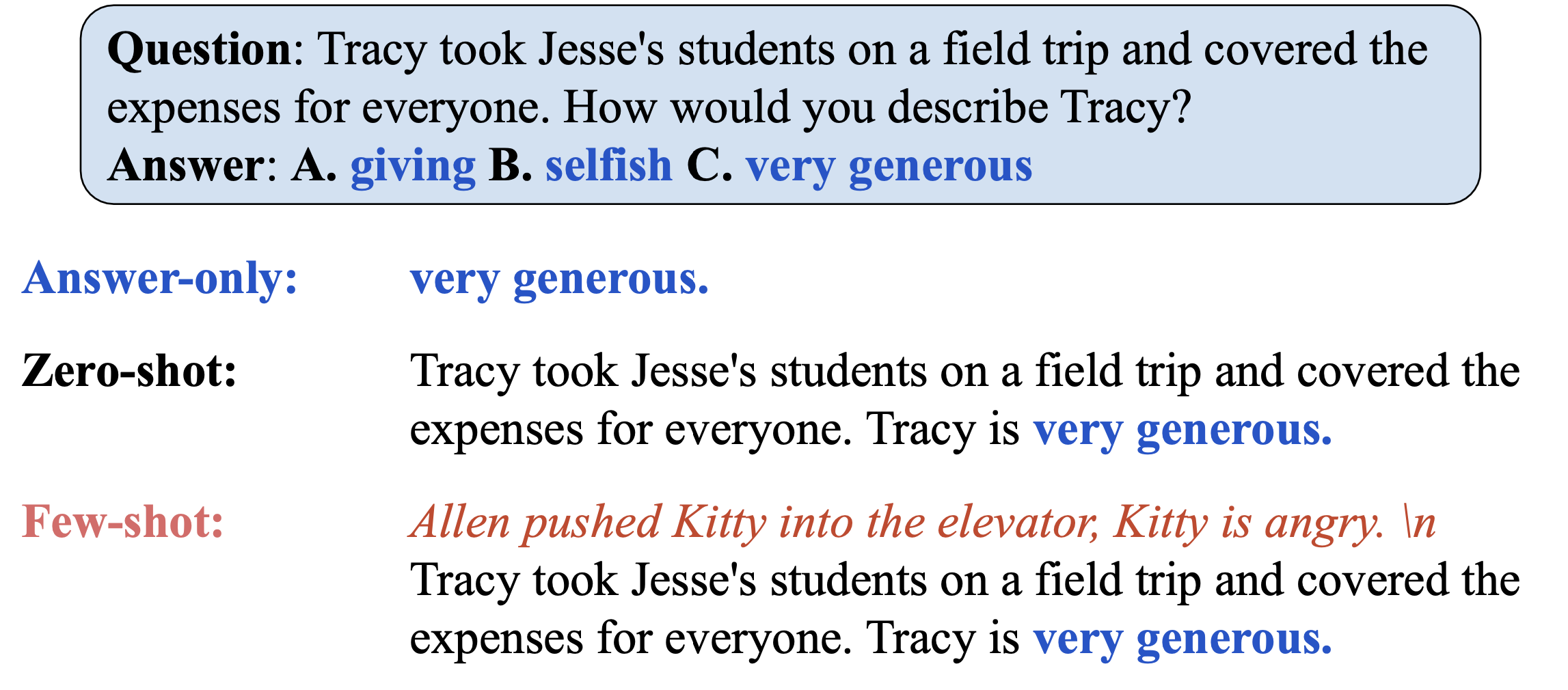}
\caption{The experiment settings with their corresponding input to the LM. The example is taken from \socialiqa{} \citep{socialiqa} where we convert questions to natural text using the rules of \citet{self_talk_2020}; this conversion yields to better performance (\S\ref{sec:robustness}).}
\label{fig:figure1}
\end{figure}

Concretely, our work differs from prior work on commonsense evaluation of LMs \citep{gpt3,mt-nlg} by way of a more rigorous evaluation, in which we: (i) carefully control for the LM's ability to exploit potential surface cues and annotation artefacts to predict the answer, without reasoning over the context. We further (ii) account for the variations in factors influencing the LM's performance, which arise from certain evaluation design choices --- independently of commonsense knowledge in the models. We systematically conduct this study on four commonsense benchmarks, six model sizes (up to a very large LM with 280B parameters), and multiple evaluation settings (\eg, different score functions and prompt format).

We begin with our first question: When evaluating a large LM in a zero-shot setting, \emph{how does its zero-shot performance compare to a strong baseline (\S\ref{sec:zero-shot}})? Controlling for the LM's ability to guess the correct answer, \emph{without} even looking at the question \citep[][\textbf{Answer-only baseline}, top of \Figure{fig:figure1}]{hypothsesis-only-nli-baselines,trichelair-etal-2019-reasonable}, we find that, despite the LM's strong zero-shot performance, the Answer-only baseline can nevertheless perform surprisingly well on some benchmarks. Despite the clear importance of comparing with answer-only baselines as shown in Figure \ref{fig:zs}, these comparisons are absent from recent work on large LMs \citep{zhou2020evaluating,gpt3,rae-etal-2021-gopher}. 
Furthermore, increasing model size alone is unlikely to bridge the gap with human performance in the near future: Our analysis of scaling behavior suggests that much larger dense LMs (with $100$T to $10^{18}$ parameters --- which are infeasibly large at present) are needed to achieve human performance for 3 out of 4 benchmarks.

\emph{Does familiarizing the LM with the task format using a few-shot evaluation setting substantially improve performance (\S\ref{sec:few-shot})?}
We find that the few-shot evaluation (using up to 64 examples) does not substantially improve the LMs' performance for most tasks except \socialiqa{}. Moreover, using the few-shot/in-context demonstration setting fails to bridge the gap between the LM and current SOTA.

Finally, we ask: \emph{to what extent does the model's zero-shot performance vary depending on certain evaluation design choices, such as the format of the prompt or the score function 
(\S\ref{sec:robustness})}? We find that these design choices --- though they have little to do with common sense --- can result in large fluctuations in performance {(up to 19\%)}.
This finding challenges the notion that large LMs are largely able to work well out-of-the-box with minimal task-specific tuning. Based on these findings, we emphasize the need to carefully select such design choices, explicitly state them to enable fair comparison with prior work, and quantify the robustness of the observed results across different design choices.

All in all, our findings suggest that acquiring \emph{human-level} commonsense knowledge, without relying on surface cues or task-specific supervision, remains beyond the reach of current large LMs. Given the marginal improvements from increasing model size, we conjecture that other techniques, such as explicit commonsense supervision, multi-modal grounding, or physical embodiment \citep{bisk-etal-2020-experience}, are promising ways forward.

\section{Experimental Setting}
\begin{table}[t]
\centering
\tiny
\begin{tabular}{@{}l|lll@{}}
\toprule
                    & \textbf{Choices} & \textbf{Knowledge Types} & \textbf{Questions} \\ \midrule
\textbf{\hellaswag} \cite{hellaswag}  & 4                          & Temporal, Physical  &  10042                        \\
\textbf{\winogrande} \cite{winogrande} & 2                          & Social, Physical    & 1267                        \\
\textbf{\socialiqa} \cite{socialiqa}  & 3                          & Social                   & 1954                         \\
\textbf{\piqa}  \cite{piqa}& 2                          & Physical                 & 1838                         \\ \bottomrule
\end{tabular}

 \caption{Benchmark Statistics. Choices: the number of candidate answers for each question; Questions: the number of candidate answers for each question.}
 \label{tab:stats}
\end{table}

We begin by outlining our experimental setup, and describe the benchmarks, model, baselines, and other relevant experimental settings.

\subsection{Commonsense Benchmarks}
\label{sec:benchmarks}

Commonsense knowledge spans many categories, such as physical common sense (\eg, a car is heavier than an apple), social common sense (\eg, a person will feel happy after receiving gifts), and temporal common sense (\eg, cooking an egg takes less time than baking a cake). Given this diverse nature of commonsense knowledge, various benchmarks have been proposed to test these different types of knowledge \citep[\eg,][]{hellaswag,winogrande,socialiqa,piqa,commongen,boratko2020protoqa}.

Commonsense benchmarks broadly consist of two tasks: (a) multiple-choice evaluation \citep[][]{zellers2018swag,hellaswag,socialiqa,piqa}, where a model needs to choose the correct answer from a list of plausible answers; (b) generative evaluation \citep{boratko2020protoqa,commongen,lin2020differentiable}, which requires a model to generate an answer given a question and some additional context. Here we focus on multiple-choice benchmarks, since they provide a more reliable automatic metric (\ie, accuracy), whereas automated metrics used to evaluate language generation \citep[\eg, BLEU,][]{Papineni2002BleuAM} do not correlate perfectly with human judgment \citep{liu-etal-2016-evaluate,novikova-etal-2017-need}.\footnote{Human judgment of LM output is not only costly to obtain, but also imperfect \cite{clark-etal-2021-thats}, compounding the difficulty of commonsense evaluation in a generation setup.} We use a diverse set of four representative multiple-choice commonsense benchmarks to better understand the extent to which \pretrained{} LMs are able to acquire different types of commonsense knowledge. We use the validation split of each benchmark, as their test splits are not public.

\noindent\textbf{\hellaswag} \citep{hellaswag} is designed to evaluate a model's ability to understand physical, grounded, and temporal common sense.  Given a four-sentence story, the model must choose the correct ending from four candidates. The stories are either video captions from AcitivityNet \cite{activity_net}, or WikiHow passages \cite{koupaee2018wikihow}. When evaluating LMs on a similar dataset~\citep{zellers2018swag}, incorrect answers can be easy to distinguish from correct ones; hence in constructing \hellaswag, \citet{hellaswag} removed easy negatives through adversarial filtering.

\begin{table*}[]
\centering
\small
\begin{tabular}{@{}l|l|l@{}}
\toprule
\textbf{Dataset}           & \multicolumn{1}{c|}{\textbf{Prompt: $x$}}                                                                                                          & \multicolumn{1}{c}{\textbf{Answer: $y$}}                                                                      \\ \midrule
\textbf{\hellaswag}  & \begin{tabular}[c]{@{}l@{}}A woman is outside with a bucket and a dog. The dog is running \\ around trying to avoid a bath. She\end{tabular}      & gets the dog wet, then it runs away again.                                                                  \\ \midrule
\textbf{\winogrande} & The GPS and map helped me navigate home. I got lost when the                                                                             &  \textbf{GPS} got turned off.                                                                                                  \\ \midrule
\textbf{\socialiqa}  & \begin{tabular}[c]{@{}l@{}}Jordan was in charge of taking the food on the camping trip and \\ left all the food at home. Jordan felt\end{tabular} & \begin{tabular}[c]{@{}l@{}}horrible that he let his friends down on\\ the camping trip.\end{tabular}        \\ \midrule
\textbf{\piqa}       & Make Halloween lanterns.                                                                                                                         & \begin{tabular}[c]{@{}l@{}}Draw ghost faces on empty milk bottles,\\ put a candle in each one.\end{tabular} \\ \bottomrule
\end{tabular}
 \caption{Examples of the prompt $x$ and the correct answer $y$ in different benchmarks.} 
 \label{tab:data_example}
\end{table*}

\noindent\textbf{\winogrande} \cite{winogrande} is a co-reference resolution benchmark that mainly examines physical and social common sense. Each example consists of a sentence (\eg, ``The trophy did not fit the suitcase because it is too big.'') and two candidate \emph{entities} (\eg, ``trophy'' or ``suitcase''). The task is to choose the correct entity for the pronoun, \eg, ``it'' refers to ``trophy'' in the example. 

\noindent\textbf{\socialiqa} \cite{socialiqa} focuses on evaluating social commonsense, in particular theory of mind --- the capacity to reason about others' mental states \cite{flavell2004theory}. 
Given context sentences and a corresponding question, the task is to choose the correct response from three candidates. Annotators use the ATOMIC knowledge base \cite{sap2019atomic} to create context sentence and questions; the answers are provided by additional annotators.  

\noindent\textbf{\piqa} \cite{piqa}, short for physical interaction question answering, mainly covers 
 the physical aspect of common sense.  
Each data point
consists of a 
task and two alternative solutions to finish the task; one of which is correct. The tasks are curated from a website\footnote{\url{https://www.instructables.com/}} with instructions for everyday tasks (\eg, separating egg yolks from eggs); the solutions are provided by human annotators.

\subsection{Pre-trained Language Model}
\label{sec:language_model}

We use the \pretrained{} language model of \citet{rae-etal-2021-gopher}, Gopher, which is an autoregressive Transformer \citep{vaswani2017attention} language model with 280 billion parameters. We choose Gopher because of its excellent zero-shot and few-shot performance at various benchmarks, in addition to its large model size, which has been shown to improve language modeling and downstream performance \citep{kaplan_2020}. Notably, Gopher is more than 50\% larger than GPT3 and as of March 2022, is one of the largest dense LMs developed to date.
\paragraph{Gopher hyper-parameters.} The \pretrained{} Gopher language model has 80 layers, 128 attention heads, 128-dimensional key/value vectors, and a feedforward layer dimension of 16,384. To better understand the effect of different model sizes (\S\ref{sec:model_size_analysis}), we experiment with five other model sizes: 44M, 117M, 417M, 1.4B, and 7.1B. Similar to Gopher, each of these models was \pretrained{} by \citet{rae-etal-2021-gopher}; a full list of model hyper-parameters is summarized in Table 1 of \citet{rae-etal-2021-gopher}. Each model is trained by subsampling from the MassiveText dataset, which consists of more than 2 trillion tokens from various domains including web pages, news, books, and codes \citep{rae-etal-2021-gopher}. The authors have removed documents that overlap significantly with the evaluation sets from training set including  benchmarks used in our work.
We use TPUv3 to conduct all evaluations, with an estimated total compute budget of $2 \times 10^{20}$ FLOPs.

\paragraph{Score function.} On the multiple-choice benchmarks, we evaluate the \pretrained{} LM by calculating the score for each answer choice under the model, and select the highest-scoring answer $\bold{\hat{y}}$:
\begin{equation*}
\label{eq: multiple_question_score}
    \bold{\hat{y}} = \operatorname*{arg\,max}_{\bold{y} \in Y(\bold{x})} s_{\boldsymbol{\theta}}(\bold{y}|\bold{x});
\end{equation*}
here $\bold{x}$ denotes the question or prompt, $Y(\bold{x})$ the set of answer choices for a given question, and $s_{\boldsymbol{\theta}}(\cdot)$ 
the score of an answer choice $\bold{y}$ given $\bold{x}$, under the \pretrained{} LM with parameters $\boldsymbol{\theta}$. 
We provide some examples 
in \Table{tab:data_example}.\footnote{For \socialiqa{}, we concatenate the context sentence and question together to form the prompt $\bold{x}$.} For \socialiqa{}, we convert questions to natural text using the rules of \citet{self_talk_2020}; we find this natural text format to yield better results, as discussed in \Section{sec:robustness}.

Unless otherwise stated, we use \emph{cross-entropy} (or token-level log prob) to score each answer:
\begin{equation}
    s_{\boldsymbol{\theta}}(\bold{y}|\bold{x}) = \frac{\sum_{i =0} ^{\|\bold{y}\|}\textit{log}(p_{\boldsymbol{\theta}}(y_i|x,y_{0}...y_{i-1}))}{\|\bold{y}\|}. \label{eq:score_function}
\end{equation}
This score function reduces the impact of length; without dividing by $\|\bold{y}\|$, longer answers might have lower probabilities \cite{stahlberg-byrne-2019-nmt}.
GPT3 \citep{gpt3} also employs this score function for zero-shot evaluation.

\subsection{Baselines}\label{sec:baselines}
We compare the performance of Gopher with two baselines. The first, simple baseline is to randomly select an answer candidate, where the chance of selecting the correct one is  $\frac{1}{\textit{number of choices}}$. We henceforth refer to this as the \emph{Random Baseline}. We experiment with two other baselines: Either choosing the majority label from the training data, or choosing the longest answer. We omit these baselines as they perform similarly to the Random Baseline.

More importantly, we consider an \emph{Answer-only Baseline}, where we select the highest-scoring answer choice under the LM, \emph{without} conditioning on the question. More formally, this baseline considers $s_{\boldsymbol{\theta}}(\bold{y})$, as opposed to $s_{\boldsymbol{\theta}}(\bold{y} | \bold{x})$ in Eq.~\ref{eq:score_function}. This baseline reveals the extent to which the \pretrained{} LM conducts the appropriate reasoning over the context to select the answer, as opposed to relying on potential surface cues or annotation artefacts that make the correct answer \emph{a priori} more probable than the rest. We illustrate this baseline at the top of \Figure{fig:figure1}. For \winogrande{}, we calculate the cross-entropy of the text starting by the pronoun replacement, as shown in \Table{tab:data_example}. Ideally, each answer choice should be equally likely if we do not consider the question, and the Answer-only performance should be close to the Random baseline. Similar hypothesis-only baselines are well-studied for natural language inference datasets \cite{hypothsesis-only-nli-baselines}; \citet{trichelair-etal-2019-reasonable} further explored such an Answer-only baseline, albeit only on the SWAG benchmark \cite{zellers2018swag}.

\section{Zero-shot Performance}
\label{sec:zero-shot}
\begin{figure*}[]
    \centering
    \includegraphics[width=\textwidth]{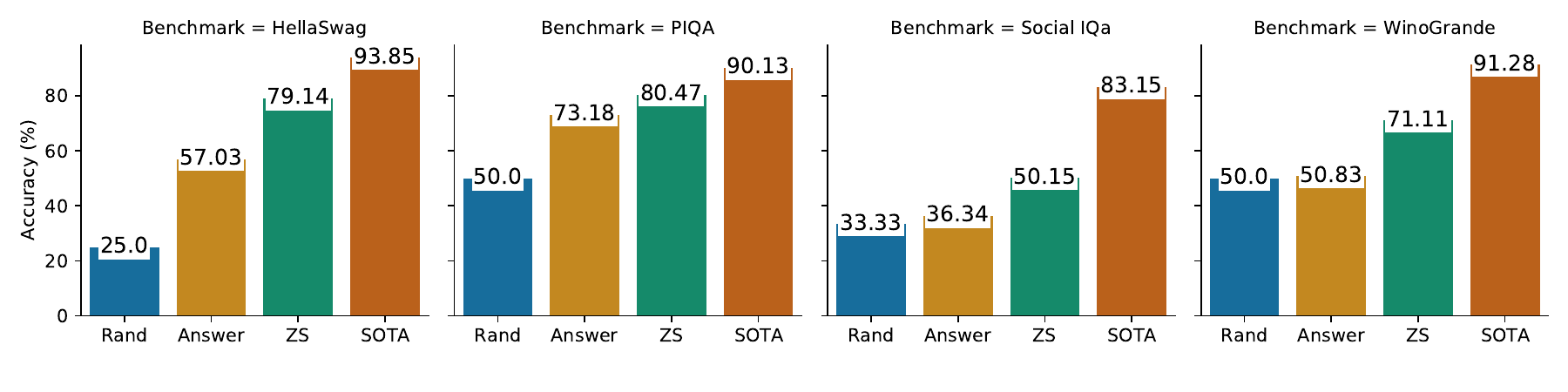}
    \caption{Random Baseline (Rand), Answer-only Baseline (Answer), zero-shot (ZS), and the current state-of-the-art (SOTA) for each benchmark, which is achieved by UNICORN \citep{Lourie2021UNICORNOR}.}
    \label{fig:zs}
\end{figure*}

In \Figure{fig:zs}, we report the zero-shot performance of our \pretrained{} LM (with 280B parameters, \S\ref{sec:language_model}) on the four commonsense benchmarks, alongside: (i) the Random and Answer-only baselines, and (ii) the current state-of-the-art (SOTA) result. The SOTA results are achieved by the UNICORN~\citep{Lourie2021UNICORNOR} model with 11B parameters, which is \pretrained{} on 6 existing commonsense datasets~\citep{hellaswag,piqa,socialiqa,winogrande,bhagavatula2020abductive,Huang2019CosmosQM}.

\paragraph{Zero-shot performance.} 
At first glance, we observe strong zero-shot results, outperforming the Random Baseline in all benchmarks (compare ``Rand'' and ``ZS'' in \Figure{fig:zs}). However, the gap between the stronger Answer-only baseline and the zero-shot result is smaller for all benchmarks  (compare ``Answer'' and ``ZS''): Whereas this gap is still sizable for \hellaswag{} and \winogrande{} (>20), it is much smaller for \socialiqa{} and \piqa{}. Finally, in all cases, there is still a large gap between the SOTA and zero-shot performance (>10); this gap is largest for \winogrande{} and \socialiqa{}, suggesting that social and physical commonsense is challenging for \pretrained{} LMs --- even a large one with 280B parameters --- without task-specific supervision.\footnote{We remark that the 530B-parameter LM of \citet{mt-nlg} achieves slightly better performance than Gopher on \hellaswag{} (80.2), \piqa{} (82), and \winogrande{} (73), although there remains a large gap with the SOTA performance.} 

\subsection{Answer-only bias} 
\begin{figure}[t]
    \centering
  \includegraphics[width=0.45\textwidth]{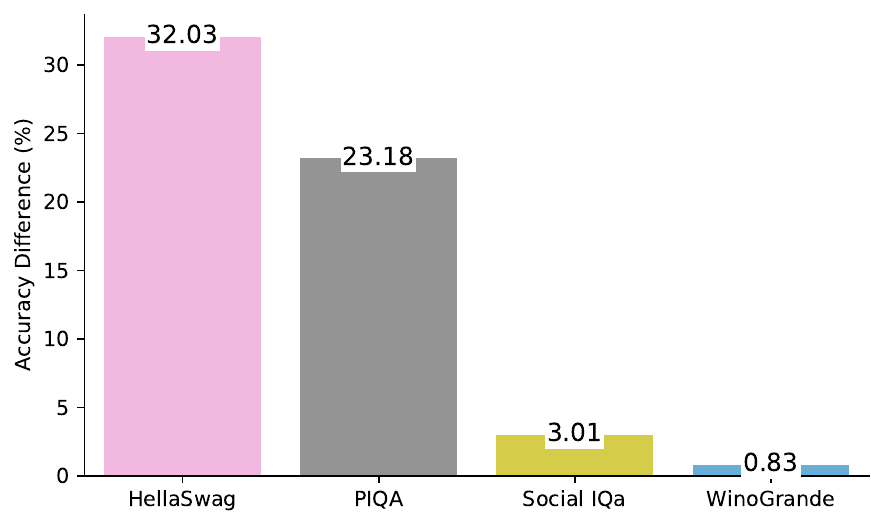}
    \caption{The performance gap between Answer-only and Random baselines for each benchmark.}
    \label{fig:zs_difference}
\end{figure}

As shown in \Figure{fig:zs_difference}, the performance gap between the Random and Answer-only baselines is notably large for \hellaswag{} and \piqa{}, where the Answer-only baseline outperforms the Random baseline by more than 32\% and 23\%, respectively. This large gap highlights an existing answer-only bias in these benchmarks: the correct answer can, in fact, be selected by the LM without conducting the appropriate commonsense reasoning over the provided context.
On the other hand, the Answer-only baseline performs similarly to the random baseline on \winogrande{} and \socialiqa{}; hence, the zero-shot performance on these benchmarks is a more reliable estimate of the model's acquisition of commonsense knowledge. 
Given the existing (and sometimes inevitable) answer-only biases in some benchmarks, it is important to contextualize the zero-shot results by comparing with strong baselines, although such comparisons are missing from recent work \citep[\eg,][]{zhou2020evaluating,gpt3,rae-etal-2021-gopher}. 
 
\subsection{Does Increasing Model Size Help?}
\label{sec:model_size_analysis}
\begin{table}[]
\centering
\tiny
\begin{tabular}{@{}l|l|ccccc@{}}
\toprule
\textbf{}                                   & \textbf{}                        & \multicolumn{1}{l}{\textbf{Answer}} & \textbf{ZS}                    & \textbf{FS(1)}                 & \textbf{FS(10)} & \textbf{FS(64)}                \\ \midrule
\multirow{5}{*}{\textbf{\hellaswag}}  & \textbf{44M}                     & 25.8                                & 28.0                           & 28.0                           & \textbf{28.1}   & 27.9                           \\
                                            & \textbf{117M}                    & 29.2                                & 33.5                           & 33.3                           & \textbf{34.0}   & 33.5                           \\
                                            & \textbf{417M}                    & 35.6                                & \textbf{44.1}                  & 43.4                           & 43.3            & 43.3                           \\
                                            & \textbf{1.4B}                    & 43.2                                & \textbf{56.7}                  & 56.4                           & 56.2            & 56.5                           \\
                                            & \textbf{7.1B}                    & 50.4                                & \textbf{69.5}                  & 67.6                           & 67.9            & 67.9                           \\
                                            & \textbf{Gopher}                  & 57.0                                & 79.1                           & 77.8                           & 79.2            & \textbf{79.3}                  \\ \midrule
\multirow{6}{*}{\textbf{\winogrande}} & \textbf{44M}    & 48.5                                & \textbf{51.3} & 51.1                           & 50.8            & 50.6                           \\
                                            & \textbf{117M}   & 50.8                                & 52.0                           & \textbf{51.9} & 50.9            & 50.8                           \\
                                            & \textbf{400M}   & 49.9                                & 52.2                           & 51.8                           & 50.8            & \textbf{52.5} \\
                                            & \textbf{1.3B}   & 49.7                                & \textbf{58.1} & 56.4                           & 56.0            & 57.3                           \\
                                            & \textbf{7B}     & 52.4                                & \textbf{64.6} & 62.1                           & 63.1            & 62.0                           \\
                                            & \textbf{Gopher} & 50.8                                & 71.1                           & 69.2                           & 71.4            & \textbf{74.6} \\ \midrule
\multirow{6}{*}{\textbf{\socialiqa}}  & \textbf{44M}    & 35.5                                & \textbf{42.0} & 41.2                           & 40.9            & 40.9                           \\
                                            & \textbf{117M}   & 36.1                                & \textbf{43.7} & 42.7                           & 42.1            & 42.2                           \\
                                            & \textbf{400M}   & 36.0                                & \textbf{45.6} & 44.5                           & 45.2            & 45.3                           \\
                                            & \textbf{1.3B}   & 35.8                                & 46.9                           & 46.4                           & 48.6            & \textbf{50.5} \\
                                            & \textbf{7B}     & 36.9                                & 48.1                           & 48.1                           & 52.9            & \textbf{54.2} \\
                                            & \textbf{Gopher} & 36.3                                & 50.2                           & 50.2                           & 55.3            & \textbf{57.5} \\ \midrule
\multirow{6}{*}{\textbf{\piqa}}       & \textbf{44M}    & 60.2                                & \textbf{62.6} & 62.1                           & 62.3            & 61.3                           \\
                                            & \textbf{117M}   & 62.1                                & \textbf{65.5} & 64.6                           & 65.1            & 65.3                           \\
                                            & \textbf{400M}   & 65.9                                & \textbf{70.9} & 68.8                           & 70.5            & 70.1                           \\
                                            & \textbf{1.3B}   & 68.4                                & 74.4                           & 73.3                           & 74.4            & \textbf{74.6} \\
                                            & \textbf{7B}     & 70.0                                & 77.4                           & 75.5                           & 77.6            & \textbf{78.1} \\
                                            & \textbf{Gopher} & 73.2                                & 80.5                           & 79.3                           & 81.4            & \textbf{81.5} \\ \bottomrule
\end{tabular}
\caption{Performance of all models across benchmarks under different experimental settings. Ans: Answer-only Baseline; ZS: zero-shot performance; FS($n$): few-shot performance where $n$ is the number of examples.}
\label{tab:perf}
\end{table}

\begin{figure*}[]
    \centering
  \includegraphics[width=\textwidth]{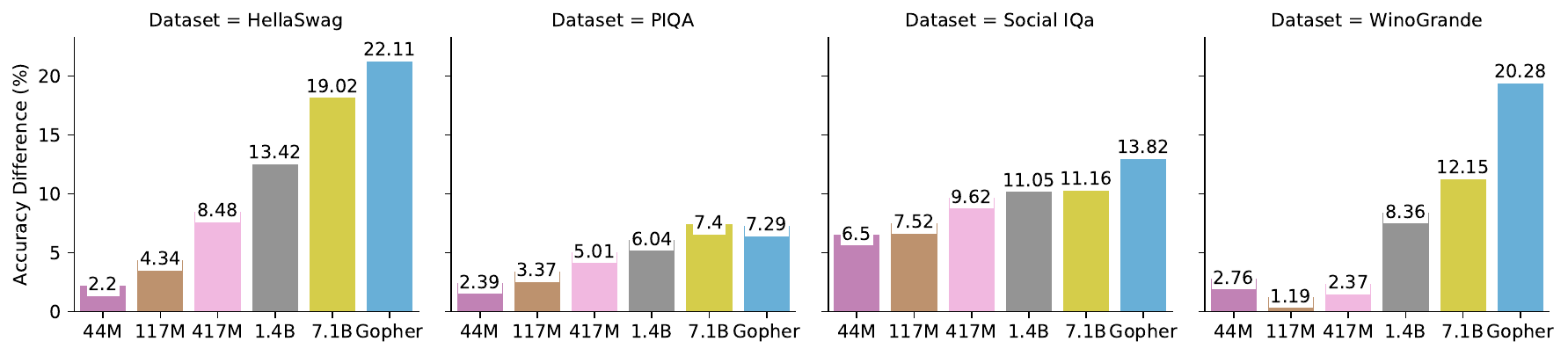}
    \caption{The difference between zero-shot performance and Answer-only baseline for different model sizes.} 
    \label{fig:answer_and_zero_difference}
\end{figure*}

Gopher (the largest LM we have access to) achieves a decent zero-shot performance for most commonsense benchmarks, but maintains a notable gap with fine-tuned SOTA results. Can we eventually reach human-level performance on these commonsense benchmarks by increasing model size alone? 

Since we do not have access to larger language models than Gopher, we examine the extent to which zero-shot performance improves when using Gopher compared to a range of smaller models (\ie, scaling plots). Such scaling plot can help us predict the performance for larger models than Gopher. To that end, we use 6 \pretrained{} model sizes from 44M to 280B parameters (see \S\ref{sec:language_model}).\footnote{Each model size is trained on the same dataset; hence any performance differences can be attributed to model size.} We present the findings in \Table{tab:perf}. On all four benchmarks, the LM's zero-shot performance (\Table{tab:perf}, \textbf{ZS} column) consistently gets better as we use increasingly larger models.
This finding is also consistent with that of \citet{gpt3}, who showed that larger models have better performance at \hellaswag{}, \winogrande{}, and \piqa{}. But, crucially, we argue that this does \emph{not} necessarily mean that larger models are better at commonsense reasoning: For \hellaswag{} and \piqa{}, the Answer-only baseline also substantially improves with model size (\Table{tab:perf}, \textbf{Answer} column). Hence, for these benchmarks, larger models are \emph{also} better at exploiting potential surface cues and annotation artefacts to guess the correct answer, without reasoning over the context. To properly assess commonsense reasoning, we should focus on the \emph{performance difference} between the zero-shot and the Answer-only baseline. 

We plot this performance difference with respect to different model sizes in \Figure{fig:answer_and_zero_difference}. We observe that larger models have better performance across benchmarks --- when increasing model size, the zero-shot performance gains are \emph{more} than the performance gains of the Answer-only baseline. Nevertheless, the magnitude of this improvement varies depending on the benchmark: We see a substantial improvement on \winogrande{}, but smaller improvements on \hellaswag{},  \socialiqa{} and \piqa{}.
\paragraph{Scaling behavior.} Based on these trends, what model size would be required to achieve human-level performance on these benchmarks? Through a linear regression analysis (see Appendix~\ref{app:scaling} for more details), given the current rate of improvement in performance when gradually increasing the model size from 44M up to 280B, we need a model of at least 1.4T parameters to achieve human performance on HellaSwag, and a model of >100T parameters ($\sim$400x larger than Gopher) for other benchmarks. This result suggests that training ever-larger models may not help us reach human performance, at least in the near future. Indeed, given the enormous compute costs for training even larger LMs than the Gopher model with 280B parameters, we conjecture that there are more efficient ways of acquiring commonsense knowledge in an unsupervised fashion, for instance through multi-modal learning and grounding \citep{bisk-etal-2020-experience}. 
\section{Few-shot Performance}
\label{sec:few-shot}

\begin{figure*}
    \centering
  \includegraphics[width=\textwidth]{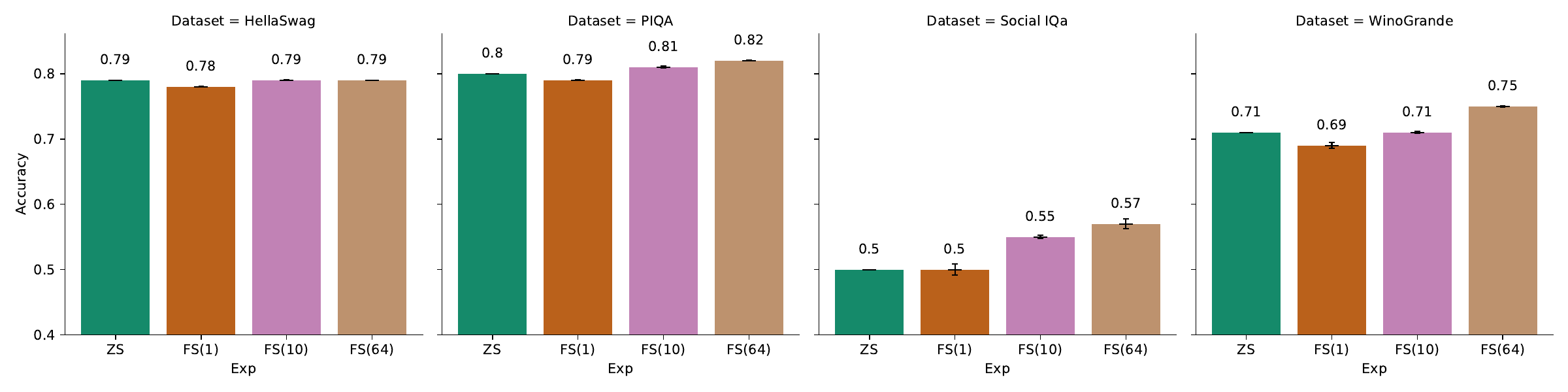}
    \caption{Accuracy on the benchmarks for zero-shot (ZS) and few-shot (FS) settings (with 1, 10, and 64 examples). We additionally report the error bars, although the error bars are not always visible due to the very small variance.}
    \label{fig:few-shot}
\end{figure*}

Recent work has shown that large LMs can perform surprisingly well at various tasks in a few-shot fashion \citep{gpt3,mt-nlg}. Under this setup, the model is provided with $n$ examples of the downstream task, which are then appended to the prefix. Concretely, for the four commonsense benchmarks, we append $n$ examples that include the question and the correct answer; these examples --- which are randomly sampled from the training split of each benchmark --- appear before the evaluated question, as shown in \Figure{fig:figure1}. 
This few-shot formulation is appealing as it relies only on a small number of task-specific examples to get the LM accustomed to the task, \emph{without} any fine-tuning. To what extent can we improve the model performance on commonsense benchmarks, by shifting from the zero-shot to the few-shot evaluation protocol?\footnote{The ability of large LMs to perform few-shot/in-context learning was first demonstrated by GPT3. Here we use an even-larger model than GPT3, which we expect to be able to leverage in-context learning to a similar extent as GPT3.}

In \Figure{fig:few-shot}, we compare the performance of Gopher under different evaluation protocols: (i) zero-shot and (ii) few-shot ($n$) where we use $n \in \{1, 10, 64\}$ examples. We run the few-shot experiments between 5 and 10 times --- sampling different examples each time --- and report the average performance. The variance across runs is very small and is shown as the error bar in \Figure{fig:few-shot}.\footnote{Our findings on the small variance with different few-shot examples is consistent with \citet{min2022rethinking}, who found that replacing real examples with random labels can work as well.} Interestingly, model performance with few-shot (1) is sometimes \emph{worse} than the zero-shot model, but the few-shot (10) and (64) models outperform their zero-shot counterpart (albeit sometimes by small margins). On \hellaswag{} and \piqa{}, we do not observe substantial improvement from few-shot evaluation compared to the zero-shot baseline (less than $2\%$).\footnote{In few-shot experiments ($n=50$), \citet{gpt3} also found small improvements for \piqa{} and \hellaswag{} ($<$1.5\%), with a larger improvement (7.5\%) for \winogrande{}.} While few-shot evaluation does not help much for most datasets, the only exception is \socialiqa{}, where the few-shot (64) model outperforms the zero-shot model by a $>7\%$ margin. We attribute this to the less natural text of \socialiqa{};\footnote{We found that Gopher has the highest perplexity when predicting \socialiqa{} answers compared to the other datasets.} hence adding task-specific examples provides information about what is expected of the task.

Overall, we observe that the usefulness of the few-shot setting is benchmark dependent. Moreover, using task-specific examples in a few-shot setting does not bridge the gap to SOTA or human performance for any of the benchmarks.

\paragraph{Knowledge base retrieval.} We further examine if adding pre-extracted commonsense knowledge base triplets to the context --- as a different form of few-shot/in-context learning --- helps improve model performance. (See Appendix~\ref{sec:kg} for details.) In contrast to work of \citet{Shwartz2020DoNL}, we observe no improvements when appending the triplets; we attribute this discrepancy to the strong performance of our base models (see \S\ref{sec:robustness}).

\section{Robustness of Reported Results}
\label{sec:robustness}

Different evaluation design choices --- such as the format of the prompt or the choice of score functions --- can impact the LM's zero-shot performance, and crucially result in different conclusions about a model's commonsense understanding ability. Moreover, the lack of a standardized zero-shot LM evaluation protocol makes direct comparisons between papers difficult \citep{self_talk_2020,bosselut2021dynamic}. To what extent can we attribute variance in the reported results to these evaluation design choices --- even though they have little to do with commonsense knowledge?
\paragraph{Model.} Quantifying the robustness of the reported results necessitates scoring a large number of examples under different evaluation design choices, which is infeasible to do with the largest (280B-parameter) model that has a slow inference speed. Hence, we conduct the following experiments using the 7B-parameter model, which is still $\sim$5 times larger than GPT2 \citep{gpt2}. 

\paragraph{Score functions.}
Prior work employs different score functions to assess the plausibility of each answer choice given a question \cite{gpt3,self_talk_2020, bosselut2021dynamic,holtzman2021surface}, which makes a direct comparison between different results challenging.
Here we investigate the impact of different score functions on the reported performance. In addition to cross-entropy (defined in \Section{sec:language_model}), we experiment with two other score functions.
The first is \emph{sequence log probability}, defined as the log probability of the answer choice $\bold{y}$ conditional on the question $\bold{x}$. Letting $y_i$ be the $i$-th token in the answer $\bold{y}$:

\begin{equation}
\label{eq: sequence_score}
    s(\bold{y}|\bold{x}) = \sum_{i =0} ^{\|\bold{y}\|}\textit{log}(p(y_i|\bold{x},y_{0}...y_{i-1}))
\end{equation}

Another widely used score function \cite{bosselut2021dynamic,holtzman2021surface} is \emph{point-wise mutual information}. This score function takes into account the probability of the answer choices alone, and the probability of the answer choices conditional on the question.
This metric assesses whether the question adds additional information, as commonsense reasoning should be established within the context of the question. 
As this score function accounts for the prior probability of answer options, it can yield lower accuracy than score functions like cross-entropy that do \emph{not} account for such factor (Answer-only baseline, \Section{sec:baselines}).
\begin{equation}
\label{eq: mutual_information}
    s(\bold{y}|\bold{x}) = PMI(\bold{y},\bold{x})=log\frac{p(\bold{y}|\bold{x})}{p(\bold{y})}
\end{equation}

\paragraph{Prompt format.}
Another important factor is the format of the prompt; here we consider a few such choices.
In addition to the concatenation of the question and the answer, we experiment with adding special symbols "[Question]" and "[Answer]" to specify the question and the answer \citep{gpt3}. Moreover, for \socialiqa{} and \piqa{}, we experiment with a set of predefined rules \cite[taken from][]{self_talk_2020} to convert the questions into sentences, which are closer to the LM's pre-training data format. 
Finally, we find that having the correct lower/upper case and punctuation is important; thus we manually checked all benchmarks to correct for case and punctuation.\footnote{Recent work learns the prefix that would maximize performance \citep[\eg,][]{li-liang-2021-prefix}. Here we focus on evaluation setups with no parameter updates, and leave this extension to future work. Our findings also indicate that the score function choice --- which is not covered by lightweight fine-tuning approaches --- is more important than the prompt format (\Section{sec:design_choice_results}).}

\paragraph{Scored text.}
The next option is whether to score the entire question--answer pair \citep{self_talk_2020}, or only the answer choice (conditional on the given question as prefix) as done by \citet{gpt3}
\ie, whether to calculate $s(\bold{x};\bold{y})$ or $s(\bold{y}|\bold{x})$, where $;$ implies text concatenation.

\subsection{Do These Design Choices Matter?}\label{sec:design_choice_results}

\Table{tab:parameters} shows the performance difference of using the worst versus the best design choices, which are independently optimized for each task. To sweep over the above design choices, instead of considering all combinations of parameters,  we iterate the options in one category (\eg, score function), while fixing the parameters in the other categories.\footnote{This decision saves compute resources, while offering a \textbf{lower bound} on the performance variations. Our goal here is not to seek the highest achievable performance, but to understand how much performance varies across different settings. }

Overall, we observe a difference between the best and worst settings on all benchmarks; this gap is especially large for \hellaswag{} and \piqa{}. This result shows that \emph{large language models do not simply work out of the box for some commonsense benchmarks}, because for some tasks, these evaluation design choices can account for a large variation in model performance. We find that the score function plays the most important role --- cross-entropy yields the highest accuracy values across most benchmarks, but sequence log probability achieves a slightly better performance for \winogrande{}. However, when using these scores, we should account for the Answer-only baseline (\Section{sec:zero-shot}). Moreover, converting questions to sentences makes the largest difference for \socialiqa{}.
We also find that scoring the answer conditional on the question --- as opposed to scoring the concatenation of questions and answers --- works best, except for \winogrande{}, which has no questions.
\begin{table}[]
\centering
\resizebox{0.35\textwidth}{!}{%
\begin{tabular}{@{}l|ccc@{}}
\toprule
                    & \textbf{Worst} & \textbf{Best} & \textbf{Difference} \\ \midrule
\textbf{\hellaswag}  & 50.8 
& \textbf{70.5}  &  19.7     \\
\textbf{\piqa}       & 62.5 
& \textbf{78.7}  &  16.2     \\
\textbf{\socialiqa}  & 43.9    
& \textbf{48.5}  &  4.6     \\
\textbf{\winogrande} & 59.7   
& \textbf{62.0}  
&  2.3  \\  \bottomrule
\end{tabular}}
\caption{ The performance difference between the worst and best design choices for each benchmark.}
\label{tab:parameters}
\end{table}

\paragraph{Answer-length bias.} Although cross-entropy generally achieves the best reported performance, this score function is sensitive to answer lengths. As shown in Appendix~\ref{app:token_level_bias}, cross-entropy tends to assign higher scores to longer answers; to varying extent, this pattern holds for \piqa{}, \socialiqa{}, and \winogrande{}. We attribute this to the higher probability assigned to subsequent tokens in the sequence, as such tokens have the most context and thus can be more easily predicted than tokens in the beginning of the answer. As longer answers have more such easier-to-predict tokens, their cross-entropy tends to be lower. This pattern is reversed in metrics such as sequence log probability, where shorter sequences often have higher scores \citep{Koehn2017SixCF,stahlberg-byrne-2019-nmt}. Note that this bias does not change the results reported in this work since there is no correlation between answer length and correctness (Appendix~\ref{app:token_level_bias}).

\paragraph{Takeaways.} We conclude this section with three concrete recommendations for future work. 
\begin{itemizesquish}
\item Although cross-entropy often achieves the best performance, it does not take into account the probability of selecting the correct answer without reasoning over the context (\S\ref{sec:zero-shot}). We recommend future work to either: (i) use cross-entropy and report the \emph{gap} with the answer-only baseline, or (ii) use the PMI score function, which \emph{already} takes the probability of the answer into account.
\item In the same way that we search for the best model hyper-parameters, future work should search over certain important evaluation design choices, such as the format of the prompt, and whether to convert the questions into declarative sentences.
\item Lastly, we strongly encourage future work to report the variance of the observed results across different design choices. This can provide an indication of the \emph{robustness} of the language models' performance on commonsense benchmarks.
\end{itemizesquish}

\section{Related Work}

While recent work evaluates LMs against commonsense benchmarks in a zero- and few-shot fashion, they do not examine the extent to which model performance can be attributed to superficial cues or annotation artefacts in a given dataset (\eg, through strong baselines), nor do they quantify how robust the model performance is under different evaluation design choices.
\citet{trichelair-etal-2019-reasonable, elazar2021back} investigate the existence of dataset bias in commonsense co-reference resolution benchmarks~\cite{levesque2012winograd,winogrande} and SWAG \cite{zellers2018swag}; here we conduct a more comprehensive investigation on four diverse commonsense benchmarks.

Another line of work probe for commonsense knowledge in LMs through knowledge base completion  \cite{Petroni2019LanguageMA, davison2019commonsense} or manually-designed probing tasks \cite{Weir2020ProbingNL, Shwartz2020DoNL}. \citet{zhou2020evaluating} evaluate \pretrained{} LMs against commonsense benchmarks and propose a new dataset requiring multi-hop reasoning.
In contrast, we focus on zero- and few-shot evaluation of commonsense understanding using the existing benchmarks.
\section{Conclusion}
We conduct a systematic and rigorous study of large LM performance on a diverse set of commonsense benchmarks, in a zero-shot and few-shot fashion. While \pretrained{} LMs can seemingly achieve a good zero-shot performance on these benchmarks, these results can be partially attributed to the LM's ability to exploit potential surface cues and annotation artefacts to guess the correct answer, without reasoning over the provided context. We further observed that substantially increasing model size yields rather small improvements on most commonsense benchmarks: Based on the scaling plots, achieving human-level performance requires much larger model sizes than what is currently feasible. In addition, model performance can be highly sensitive to certain evaluation design choices. Overall, our findings offer valuable insights and best practices for rigorously evaluating large LMs.
\section*{Ethical Considerations}
The primary aim of this paper is to conduct a systematic and rigorous commonsense evaluation of a large language model, which --- in the case of this work --- is achieved by using the \pretrained{} Gopher language model \citep{rae-etal-2021-gopher} with 280B parameters. Hence, the same risks stemming from large language model research are also broadly applicable to this work \citep{bender_2021}. We briefly discuss these ethical considerations below.

\paragraph{Training compute.} In practice, pre-training large language models like Gopher requires an enormous amount of compute, which may contribute to increased carbon emissions \citep{strubell-etal-2019-energy,patterson_2021}. In this work, we do not pre-train the language model from scratch, although we acknowledge that conducting inference and evaluation with large language models like Gopher still has substantial computational costs. Given the need to construct even-larger language models ($>$100 trillion parameters) to achieve human-level performance on most of these benchmarks in an unsupervised fashion (\S\ref{sec:model_size_analysis}), we encourage future work to focus on potentially more efficient ways of acquiring commonsense knowledge directly from data, \eg, through multi-modal learning, grounding, and human interaction \citep{bisk-etal-2020-experience}.

\paragraph{Fairness and bias.} Given the enormous size of the pre-training data --- about 2 trillion tokens in the case of Gopher pre-training --- it is conceivable that the training dataset may inadvertently contain toxic and biased material. Such toxic material --- which is not always easily identifiable in the large training dataset --- can in turn encourage the model to produce biased, harmful, or toxic output, especially when they are prompted with toxic text \citep{gehman-etal-2020-realtoxicityprompts}. In fact, \citet{rae-etal-2021-gopher} demonstrated that --- up to a certain model size --- larger language models may respond to toxic prompts with greater toxicity compared to smaller ones. Furthermore, the enormous size of the training data does not necessarily guarantee diversity: We expect the training data to contain a smaller proportion of vernacular or regional English that is used by underrepresented communities \citep{blodgett-etal-2016-demographic,bender_2021}. Furthermore, the language model may also acquire harmful biases and stereotypes, \eg, assign lower probabilities to women becoming doctors as opposed to men \citep{rudinger-etal-2018-gender,cao-daume-iii-2021-toward}. 

\paragraph{Language model misuse.} Our work highlights both the success and limitations of large language models at multiple commonsense benchmarks. Nevertheless, the success and expressive power of large language models come at the expense of potential misuse. Given their ability to generate realistic-looking --- albeit not necessarily factual --- content, large language models can also be used for malicious purposes. For instance, large language models can be used to generate convincing fake news \citep{zellers_2019}, and more powerful generator can in turn generate even more convincing and influential fake news. Given the difficulty of manually distinguishing between human-generated text and machine-generated ones \citep{clark-etal-2021-thats}, how we can better detect and defend against malicious use of large language models is an important and exciting avenue for future work.
\section*{Limitations}
There are limitations to this work: first, we only assessed models' performance on multiple-choice questions (and not in a generative setting). 
Multiple choice problems have a more reliable automatic metric; in contrast, metrics used for generative tasks do not always accurately reflect human judgment~\cite{clark-etal-2021-thats}
Second, we only evaluate the benchmarks on one family of models, the Gopher models and their variants; given the computational cost and also the lack of availability of different large language models (LLM), we cannot run our experiments on different model families than Gopher. However, we include zero-shot results on common-sense benchmarks from existing work on other LLMs in the paper (such as the GPT2 result in Table 7). Moreover, LLMs behave very similarly on various benchmarks, and we expect our results to generalize to other LLMs as well.
Last but not least, we only evaluate models that are solely trained on language. Recent multimodal models have shown impressive performance on a range of tasks~\cite{saharia2022photorealistic}. Will models trained on multiple modalities have more commonsense? We aim to answer this question in future work.
\section*{Acknowledgments}
We would like to thank Ivana Kajić, Laura Rimell for their detailed comments on our paper. Also, thanks to Stella Biderman and the anonymous reviewers for their helpful feedback. We also thank Jack W. Rae and the other authors from the Gopher paper for providing efficient evaluation pipelines for models from the Gopher family.

\typeout{}
\bibliography{tacl2018}

\newpage
\appendix
\section{Appendix Structure}
We begin by quantifying the scaling behavior of the model to predict how performance changes with larger model sizes (Appendix~\ref{app:scaling}). We then plot the relationship between cross-entropy and answer length for each of the four datasets (Appendix~\ref{app:token_level_bias}). After that, we describe experiments that use knowledge base triplets as a form of in-context learning (Appendix~\ref{sec:kg}). Lastly, in Appendix~\ref{app:examples}, we provide qualitative examples that show which examples: (i) all model sizes get right, (ii) all model sizes get wrong, and (iii) only the larger models get right.

\section{Scaling Behavior}
\label{app:scaling}
When we estimate the performance needed to reach human-level performance, we fit a linear model to estimate accuracy from $\log(\text{params})$. We derive the human performance from each respective paper and/or leaderboard.
For \hellaswag{} and \piqa{}, human-level performance is at 95\%. For \winogrande{}, it is at 94\% and for \socialiqa{} it is at 84\%. 
On \hellaswag{}, we predict that 1.4T parameters are needed to achieve human-level performance; on \piqa{} we predict 102T parameters; on \winogrande{} we predict over 2000 Trillion parameters. \socialiqa{} scales particularly poorly, and we estimate over $10^{18}$ parameters being needed.

\clearpage
\section{Cross-entropy vs answer length for all datasets}
\label{app:token_level_bias}

\begin{table}[!ht]
\begin{tabular}{cc}
\begin{subfigure}{0.45\textwidth} 
\centering
        \includegraphics[width=0.9\textwidth]{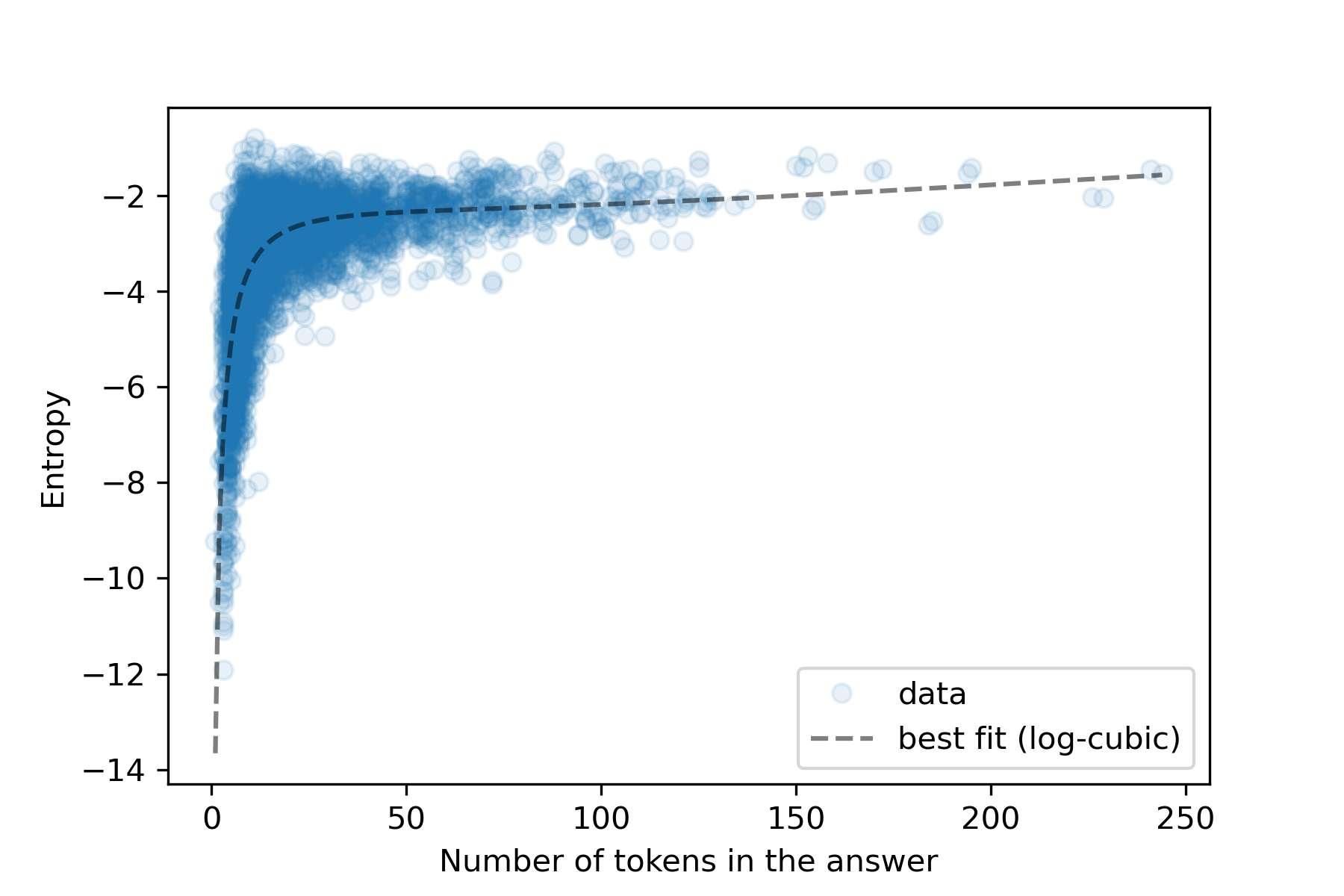}
        \caption{Answer length vs cross-entropy (average log probability across tokens) for PIQA.}
        \end{subfigure}&
\begin{subfigure}{0.45\textwidth}
        \centering
        \includegraphics[width=0.9\textwidth]{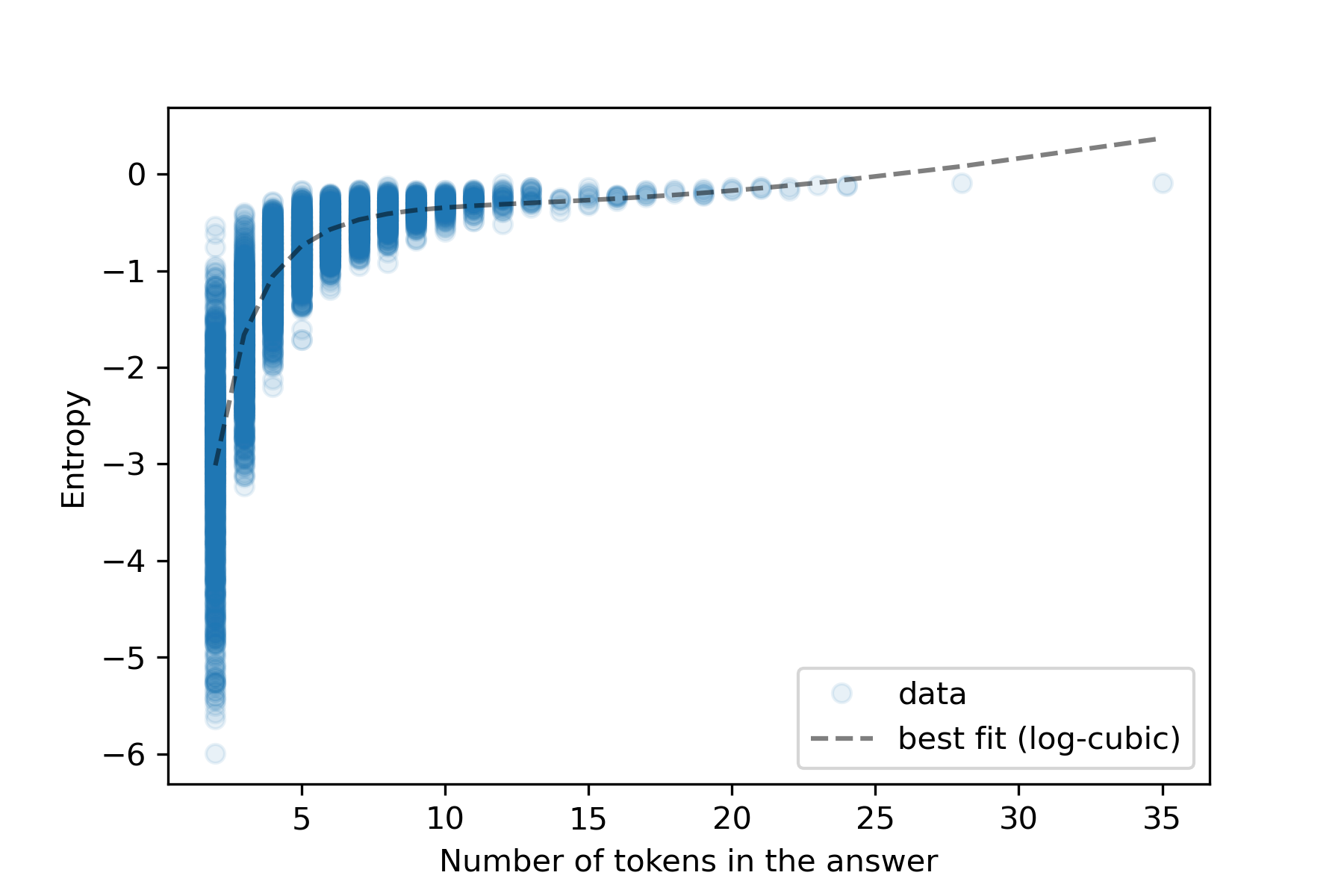}
        \caption{Answer length vs cross-entropy (average log probability across tokens) for SocialIQA.}
        \end{subfigure}
\end{tabular}
\end{table}

\begin{table}[!ht]
\begin{tabular}{cc}
\begin{subfigure}{0.45\textwidth} 
\centering
        \includegraphics[width=0.9\textwidth]{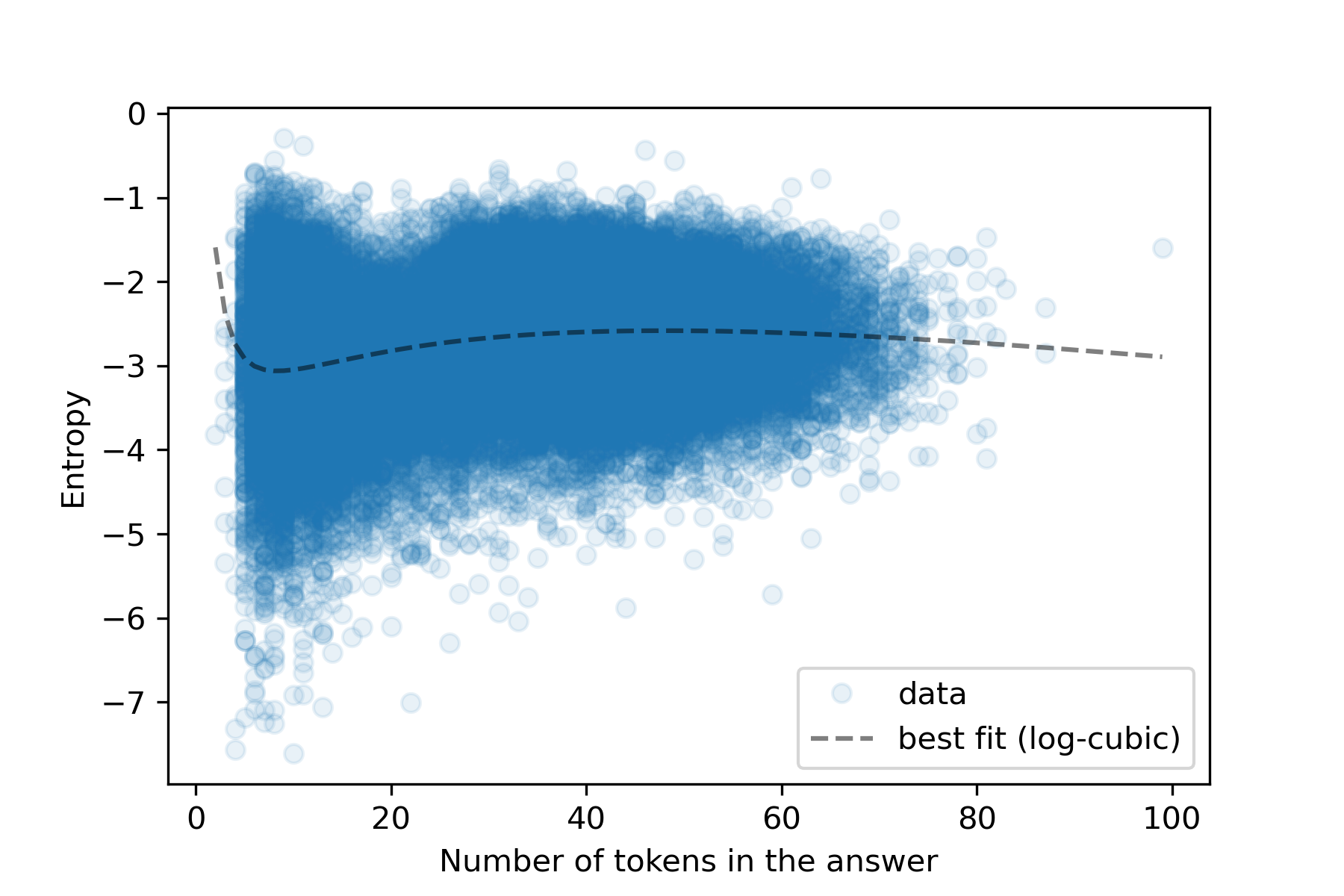}
        \caption{Answer length vs cross-entropy (average log probability across tokens) for HellaSWAG.}
        \end{subfigure}&
\begin{subfigure}{0.45\textwidth}
        \centering
        \includegraphics[width=0.9\textwidth]{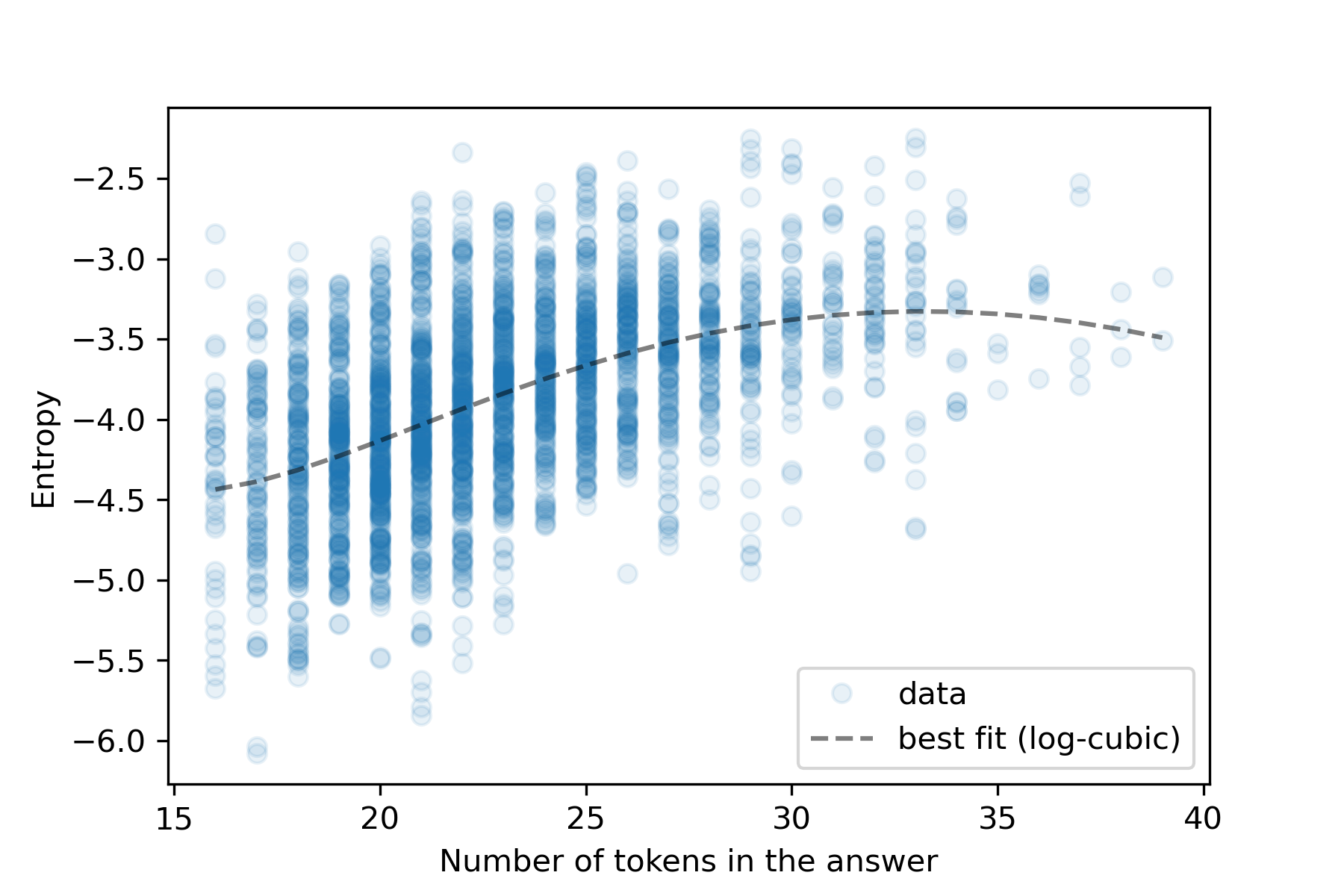}
        \caption{Answer length vs cross-entropy (average log probability across tokens) for Winogrande.}
        \end{subfigure}
\end{tabular}
\end{table}

\clearpage

\section{Commonsense Knowledge Bases}
\label{sec:kg}
Given the implicit nature of commonsense knowledge, a language model's pretraining corpora might not contain all of the supporting evidence that is required to answer commonsense understanding questions --- a phenomenon widely known as the reporting bias problem \cite{gordon2013reporting}. Thus, prior work has proposed to use external knowledge bases for improving the zero-shot performance of LMs on commonsense benchmarks \cite{bosselut2021dynamic, bauer2021identify}. 
These approaches are particularly interesting, as the knowledge base augmentation only happens at test time, rendering
this approach compatible with \emph{any} pretrained generative LM.
While prior work has shown the effectiveness of this approach over the zero-shot baseline that lacks access to commonsense knowledge bases (CSKBs), we find that the performance of the baseline model is highly sensitive to certain evaluation design choices (\S\ref{sec:robustness}). A natural question, therefore, is the following:
If we carefully optimize the evaluation design choices of the baseline model, would we \emph{still} observe similar improvements through CSKB augmentation?

\paragraph{Setup.} 
To answer this, we replicate prior work by adding commonsense knowledge base entries at test time; such knowledge base triplets can potentially provide the relevant implicit commonsense knowledge that makes the correct answer more likely than the rest. To ensure the generality of our findings, we apply this approach to multiple model sizes that we explored in \S\ref{sec:model_size_analysis}. Here we consider the pre-extracted knowledge base triplets that are made publicly available by \citet{self_talk_2020}.
We use a similar score function as \citet{self_talk_2020}, where, for each answer choice $\bold{y} \in Y(\bold{x})$, we choose the knowledge base triplet that yields the highest score:\footnote{We experimented with other score functions, such as appending the extracted knowledge base triplets to the question instead of the answer, although this approach does not yield better results than the one proposed by \citet{self_talk_2020}.}
\begin{align*}
    s_{kg}(\bold{y} | \bold{x}) &\triangleq \sum_{\bold{t}\in T} \, s(\bold{y};\bold{t} | \bold{x}) \approx \textit{max}_{\bold{t}\in T} \, s(\bold{y};\bold{t} | \bold{x}),
\end{align*}
where $s(\bold{y};\bold{t} | \bold{x})$ denotes the cross-entropy of the concatenated answer choice $\bold{y}$ and the extracted knowledge base triplet $\bold{t}$, conditional on the question/context $\bold{x}$. Here $T$ denotes the set of all extracted commonsense knowledge triplets, which are generated from Comet~\citep{Bosselut2019COMETCT}.
One key difference is that we score the answer and knowledge base triplet conditional on the question, whereas \citet{self_talk_2020} scored the concatenation of question, answer, and triplet instead.

\begin{savenotes}
\begin{table}[t!]
\centering
\small
\begin{tabular}{@{}l|lccc@{}}
\toprule
              & \textbf{ZS} & \multicolumn{1}{l}{\textbf{w/t Comet}} & \multicolumn{1}{l}{\textbf{w/t Atomic}} & \multicolumn{1}{l}{\textbf{w/t CN}} \\ \midrule
\textbf{44M}  & 42.3              & \textbf{42.9}                                  & 42.3                                   & 40.6                                       \\
\textbf{117M} & 43.6              & \textbf{44.0}                                  & 43.6                                  & 42.2                                       \\
\textbf{400M} & 46.3              & \textbf{46.8}                                  & 44.7                                   & 44.1                                       \\
\textbf{1.3B} & \textbf{47.0}              & 46.8                                  & 46.4                                   & 44.7                                       \\
\textbf{7B}   & 48.5              & \textbf{48.6}                                  & 47.5                                  & 46.1                                       \\ \midrule
\textbf{}     & \textbf{ZS} & \textbf{w/t Comet}                     & \textbf{Self-Talk}                      &                                             \\ \midrule
\textbf{GPT2} & 41.1\footnote{By similarly tuning the evaluation design choices, we achieved 46.7 when evaluating GPT2 in the zero-shot setting.}               & 47.5                                   & 46.2                                    &                                             \\ \bottomrule

\end{tabular}
 \caption{Zero-shot performance on \socialiqa{} when using different knowledge bases.  GPT2 results are taken from \citet{self_talk_2020}. ZS: zero-shot performance; CN: ConceptNet. We do not include the Gopher results --- with 280B parameters --- due to computational considerations and much slower inference.}
 \label{tab:kg}
\end{table}
\end{savenotes}

In \Table{tab:kg}, we summarize our results on \socialiqa, which has the highest gap between the zero-shot and SOTA performance (\Figure{fig:zs}).
We compare our results with those of \citet{self_talk_2020}, who used GPT2 as the base model. Our results in Table~\ref{tab:kg} provide an interesting contrast to the findings of \citet{self_talk_2020}: Our baseline zero-shot model with 1.3B parameters achieves an accuracy of 47.0\% on \socialiqa, substantially outperforming the reported GPT2 result of \citet{self_talk_2020} --- which achieves 41.1\% --- despite the fact that GPT2 has more parameters (1.5B vs our 1.3B). In fact, the same 1.3B zero-shot model
 --- which does not benefit from any commonsense knowledge base triplets --- nearly matches the performance of GPT2 augmented with Comet \citep{Bosselut2019COMETCT} (47.0\% for our zero-shot 1.3B model vs 47.5\% for GPT2 augmented with COMET; \Table{tab:kg}), and also outperforms the GPT2 model that is augmented with self-talk. 
Nevertheless, we find that adding knowledge base triplets fails to yield substantial improvements for our models; this finding is consistent across three different knowledge bases and five model sizes. On the contrary, adding such knowledge base triplets can occasionally decrease performance compared to the zero-shot baseline.

We remark on two significant aspects of our findings. First, it is important to compare proposed improvements against strong, well-tuned baselines \citep{henderson_deeprl,melis_lstm_2018}, which can achieve surprisingly competitive performance. We identify the choice of the scored span as a particularly important design choice: Whereas \citet{self_talk_2020} scored the GPT2 model on the concatenation of both question and answer, we instead calculate the cross-entropy of the answer given the question. Second, certain improvements that are observed under a particular set of evaluation design choices may not necessarily be replicated under a different set.
This finding reiterates the importance of explicitly stating the evaluation design choices used in each experiment, and identifying whether or not the observed improvements are robust across different {evaluation design choices (\Section{sec:robustness}).}

\section{Examples}
\label{app:examples}

\subsection{\socialiqa}
\textbf{All Models Incorrect}
\begin{lstlisting}[breaklines]
{'context': "Tracy didn't go home that evening and resisted Riley's attacks.",
'question': 'What does Tracy need to do before this?',
'answerA': 'make a new plan',
'answerB': 'Go home and see Riley',
'answerC': 'Find somewhere to go',
'correct': 'C'}
\end{lstlisting}
\begin{lstlisting}[breaklines]
{'context': 'Aubrey kept the baby up at night to watch for a concussion.',
'question': 'What will happen to Aubrey?',
'answerA': "The baby fell asleep despite Aubrey's best effort",
'answerB': 'gets so sleepy but stays awake anyway',
'answerC': 'and the baby both fell asleep late in the night',
'correct': 'B'}
\end{lstlisting}

\textbf{All Models Correct}
\begin{lstlisting}[breaklines]
{'context': 'Kendall opened their mouth to speak and what came out shocked everyone.',
'question': 'How would you describe Kendall?',
'answerA': 'a very quiet person',
'answerB': 'a very passive person',
'answerC': 'a very aggressive and talkative person',
'correct': 'C'}
\end{lstlisting}
\begin{lstlisting}[breaklines]
{'context': 'Sydney went to our family farm, taking the trash with her, and set it on fire on the ground.',
'question': 'How would Sydney feel afterwards?',
'answerA': 'feeling strong',
'answerB': 'burning down',
'answerC': 'upset because the fire has gotten out of control',
'correct': 'C'}
\end{lstlisting}
\begin{lstlisting}[breaklines]
{'context': 'Robin always gets pizza on the way home from work for her family on Fridays.',
'question': 'What will Robin want to do next?',
'answerA': 'pick up the pizza',
'answerB': 'complain to the others',
'answerC': 'finish work',
'correct': 'A'}
\end{lstlisting}

\textbf{Larger Models Correct}
The 1.4B, 7.1B, and 280B model all got the following correct:
\begin{lstlisting}[breaklines]
{'context': 'Alex paid extra money to get more secret details about the game strategy.',
'question': 'What will Alex want to do next?',
'answerA': 'play the game more',
'answerB': 'ignore the advice',
'answerC': 'stop playing the video game',
'correct': 'A'}
\end{lstlisting}
The 417M, 7.1B, and 280B model all got the following correct:
\begin{lstlisting}[breaklines]
{'context': 'Kai and Skylar were good friends. Kai had finally worked up the courage to ask Skylar on a date. They gave Skylar a meaningful gift to test the waters.',
'question': 'What will Kai want to do next?',
'answerA': 'say thank you for the gift',
'answerB': 'Find out whether Skylar reciprocates the feelings',
'answerC': "Tell Skylar they'd like to just be friends",
'correct': 'B'}
\end{lstlisting}

\subsection{\winogrande}
\textbf{All Models Incorrect}
\begin{lstlisting}[breaklines]
{'label': 1,
 'option1': 'Tanya',
 'option2': 'Sarah',
 'sentence': 'Tanya was unrecognizable after Sarah was done beating them, so _ ended up going to jail.'}
\end{lstlisting}
\begin{lstlisting}[breaklines]
{'label': 1,
 'option1': 'Logan',
 'option2': 'Justin',
 'sentence': 'After Logan pitched a ball that got clobbered for a home run by Justin in a baseball game, _ felt exultant.'}
\end{lstlisting}

\textbf{All Models Correct}
\begin{lstlisting}[breaklines]
{'label': 1,
 'option1': 'sausage',
 'option2': 'ball',
 'sentence':b'When the dog behaves I like to give him a sausage otherwise I give him a ball. I gave him the _ since he was bad.'}
\end{lstlisting}
\begin{lstlisting}[breaklines]
{'label': 1,
 'option1': 'Kayla',
 'option2': 'Natalie',
 'sentence': 'Kayla always wears sunscreen outdoors but Natalie doesn't because _ isn't concerned about getting neck wrinkles.'}
\end{lstlisting}

\textbf{Only Large Models Correct}
Models 400M and larger got the following correct:
\begin{lstlisting}[breaklines]
{'label': 0,
 'option1': 'Nick',
 'option2': 'Ryan',
 'sentence': 'Nick did not like sauces made from tomato, only creamy sauces. Ryan knew this so he only made white sauce when _ came over.'}
\end{lstlisting}
Models 1.4B and larger got the following correct:
\begin{lstlisting}[breaklines]
{'label': 0,
 'option1': 'Adam',
 'option2': 'Jason',
 'sentence': 'Adam loved dogs but Jason was afraid of them, so only _ petted the poodle.'}
\end{lstlisting}

\end{document}